\documentclass{article}

\usepackage[nonatbib, final]{nips_2018}

\usepackage[utf8]{inputenc} % allow utf-8 input
\usepackage[T1]{fontenc}    % use 8-bit T1 fonts
\usepackage{url}            % simple URL typesetting
\usepackage{booktabs}       % professional-quality tables
\usepackage{amsfonts}       % blackboard math symbols
\usepackage{nicefrac}       % compact symbols for 1/2, etc.
\usepackage{microtype}      % microtypography

\usepackage{cleveref}
\usepackage{graphicx}       % for pdf, bitmapped graphics files
\usepackage{amsmath}        % assumes amsmath package installed
\usepackage{amssymb}        % assumes amsmath package installed
\usepackage{lettrine}
\usepackage{adjustbox}
\usepackage{multirow}
\usepackage[bookmarks=false]{hyperref} % hyperlinks
\hypersetup{
colorlinks,
citecolor=black,
filecolor=black,
linkcolor=black,
urlcolor=black,
}

\title{Multi-Task Graph Autoencoders}

\author{
  Phi Vu Tran\\
  Strategic Innovation Group\\
  Booz $\vert$ Allen $\vert$ Hamilton\\
  San Diego, CA USA\\
  \texttt{tran\_phi@bah.com}\\
}

\begin{document}

\maketitle

%\begin{abstract}
%We examine two fundamental tasks associated with graph representation learning: link prediction and node classification. We present a new autoencoder architecture capable of learning a joint representation of local graph structure and available node features for the simultaneous multi-task learning of unsupervised link prediction and semi-supervised node classification. Our simple, yet effective and versatile model is efficiently trained end-to-end in a single learning stage, whereas previous related deep graph embedding methods require multiple training steps that are difficult to optimize. We provide an empirical evaluation of our model on five benchmark relational, graph-structured datasets and demonstrate significant improvement over three strong baselines for graph representation learning. Reference code and data are available at https://github.com/vuptran/graph-representation-learning
%\end{abstract}

\section{Autoencoder Architecture for Link Prediction and Node Classification}

As the world is becoming increasingly interconnected, relational data are also growing in ubiquity. In this work, we examine the task of learning to make predictions on graphs for a broad range of real-world applications. Specifically, we study two canonical subtasks associated with relational, graph-structured datasets: link prediction and node classification (LPNC). A graph is a partially observed set of edges and nodes (or vertices), and the learning task is to predict the labels for edges and nodes. In real-world applications, the input graph is a network with nodes representing unique entities and edges representing relationships (or links) between entities. Further, the labels of nodes and edges in a graph are often correlated, exhibiting complex relational structures that violate the general assumption of independent and identical distribution fundamental in traditional machine learning \cite{Hassan:2010}. Therefore, models capable of exploiting topological structures of graphs have been shown to achieve superior predictive performances on many LPNC tasks \cite{Rossi:2012}.

\begin{figure}[hb]
\centering
\includegraphics[width=\textwidth]{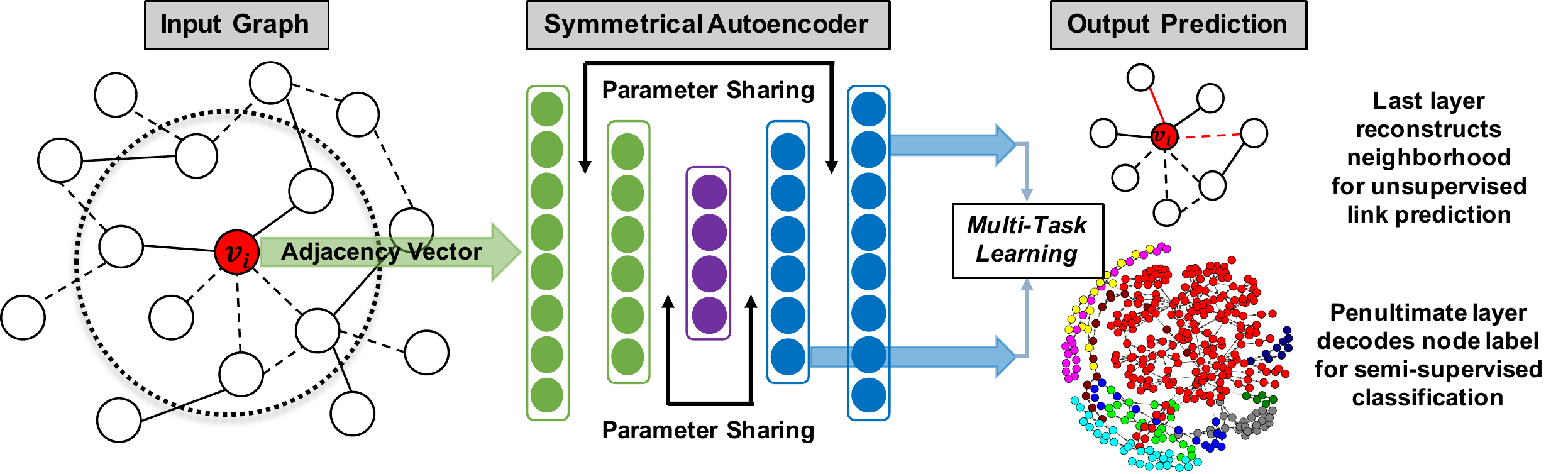}
\centering
\caption{Schematic depiction of the Multi-Task Graph Autoencoder (MTGAE) architecture. \emph{Left}: A partially observed graph with positive links (solid lines) and negative links (dashed lines) between two nodes; pairs of nodes not yet connected have unknown status links. \emph{Middle}: A symmetrical, densely connected autoencoder with parameter sharing is trained end-to-end to learn node embeddings from the adjacency vector. \emph{Right}: Exemplar multi-task output for link prediction and node classification.}
\label{fig1}
\end{figure}

We introduce the Multi-Task Graph Autoencoder (MTGAE) architecture, schematically depicted in Figure~\ref{fig1}, capable of learning a shared representation of latent node embeddings from local graph topology and available explicit node features for LPNC. Our simple, yet effective and versatile model is efficiently trained end-to-end for the joint, simultaneous multi-task learning (MTL) of unsupervised link prediction and semi-supervised node classification in a single stage, whereas previous related deep graph embedding methods require multiple training steps that are difficult to optimize. We present an empirical evaluation of the MTGAE model on five challenging benchmark graph-structured datasets and show significant improvement in predictive performance over three strong baselines designed specifically for LPNC. Reference code and data are available at \hbox{\url{https://github.com/vuptran/graph-representation-learning}}.

\noindent \textbf{Problem Formulation and Notation} ~ The input to the MTGAE model is a graph $\mathcal{G} = (\mathcal{V}, \mathcal{E})$ of $N = |\mathcal{V}|$ nodes. Graph $\mathcal{G}$ is represented by its adjacency matrix $\mathbf{A} \in \mathbb{R}^{N \times N}$ paired with a unique ordering of vertices. For a partially observed graph, $\mathbf{A} \in \{1,0,\textsc{unk}\}^{N \times N}$, where $1$ denotes a known positive edge, $0$ denotes a known negative edge, and \textsc{unk} denotes an unknown status (missing or unobserved) edge. In general, the input to the model can be directed or undirected, weighted or unweighted, and/or bipartite graphs.

Optionally, we are given a matrix of available node features, i.e. side information $\mathbf{X} \in \mathbb{R}^{N \times F}$. The aim of the MTGAE model $h(\mathbf{A,X})$ is to learn a set of low-dimensional latent variables for the nodes $\mathbf{Z} \in \mathbb{R}^{N \times D}$ that can produce an approximate reconstruction output $\mathbf{\hat{A}}$ such that the empirical error between $\mathbf{A}$ and $\mathbf{\hat{A}}$ is minimized, thereby preserving the global graph structure. In this paper, we use capital variables (e.g., $\mathbf{A}$) to denote matrices and lower-case variables (e.g., $\mathbf{a}$) to denote row vectors. For example, we use $\mathbf{a}_i$ to mean the $i$th row of the matrix $\mathbf{A}$.

\noindent \textbf{Unsupervised Link Prediction} ~ Let $\mathbf{a}_i \in \mathbb{R}^N$ be an \emph{adjacency vector} of $\mathbf{A}$ that contains the local neighborhood of the $i$th node. Our proposed MTGAE architecture comprises a set of non-linear transformations on $\mathbf{a}_i$ summarized in two component parts: encoder $g(\mathbf{a}_i)\colon \mathbb{R}^N \to \mathbb{R}^D$ and decoder $f\left(g\left(\mathbf{a}_i\right)\right)\colon \mathbb{R}^D \to \mathbb{R}^N$. We stack two layers of the encoder part to derive $D$-dimensional latent feature representation of the $i$th node $\mathbf{z}_i \in \mathbb{R}^D$, and then stack two layers of the decoder part to obtain an approximate reconstruction output $\mathbf{\hat{a}}_i \in \mathbb{R}^N$, resulting in a four-layer autoencoder architecture. Note that $\mathbf{a}_i$ is highly sparse, with up to 80 percent of the edges missing at random in some of our experiments, and the dense reconstructed output $\mathbf{\hat{a}}_i$ contains the predictions for the missing edges. The hidden representations for the encoder and decoder parts are computed as follows:
\begin{align*}
\text{Encoder} \qquad \mathbf{z}_i &= g\left(\mathbf{a}_i\right) = \text{ReLU}\left(\mathbf{W} \boldsymbol{\cdot} \text{ReLU}\left(\mathbf{V}\mathbf{a}_i + \mathbf{b}^{(1)}\right) + \mathbf{b}^{(2)}\right). \\
\text{Decoder} \qquad \mathbf{\hat{a}}_i &= f\left(\mathbf{z}_i\right) = \mathbf{V}^\text{T} \boldsymbol{\cdot} \text{ReLU}\left(\mathbf{W}^\text{T}\mathbf{z}_i + \mathbf{b}^{(3)}\right) + \mathbf{b}^{(4)}. \\
\text{Autoencoder} \qquad \mathbf{\hat{a}}_i &= h\left(\mathbf{a}_i\right) = f\left(g\left(\mathbf{a}_i\right)\right).
\end{align*}
The choice of non-linear, element-wise activation function is the rectified linear unit $\text{ReLU}(\mathbf{x}) = \text{max}(0, \mathbf{x})$. The last decoder layer computes a linear transformation to score the missing links as part of the reconstruction. We constrain the MTGAE architecture to be symmetrical with shared parameters for $\{\mathbf{W},\mathbf{V}\}$ between the encoder and decoder parts, resulting in almost $2\times$ fewer parameters than an unconstrained architecture. Parameter sharing is a powerful form of regularization that helps improve learning and generalization, and is also the main motivation behind MTL \cite{Caruana:1993, Yang:2017}. Notice the bias units $\mathbf{b}$ do not share parameters, and $\left\{\mathbf{W}^\text{T}, \mathbf{V}^\text{T}\right\}$ are transposed copies of $\{\mathbf{W}, \mathbf{V}\}$. For brevity of notation, we summarize the parameters to be learned in $\theta = \left\{\mathbf{W}, \mathbf{V}, \mathbf{b}^{(k)}\right\}, k=1,...,4$.

Optionally, if a matrix of node features $\mathbf{X} \in \mathbb{R}^{N \times F}$ is available, then we concatenate $(\mathbf{A},\mathbf{X})$ to obtain an \emph{augmented} adjacency matrix $\mathbf{\bar{A}} \in \mathbb{R}^{N \times (N + F)}$ and perform the above encoder-decoder transformations on $\mathbf{\bar{a}}_i$ for unsupervised link prediction. The intuition behind the concatenation of node features is to enable a shared representation of both graph and node features throughout the autoencoding transformations by way of the tied parameters $\{\mathbf{W},\mathbf{V}\}$.

During the forward pass, or inference, the model takes as input an adjacency vector $\mathbf{a}_i$ and computes its reconstructed output $\mathbf{\hat{a}}_i = h(\mathbf{a}_i)$ for unsupervised link prediction. During the backward pass, we learn $\theta$ by minimizing the Masked Balanced Cross-Entropy (MBCE) loss, which allows only the contributions of those parameters associated with observed edges, as in \cite{Kuchaiev:2017,Sedhain:2015}. We handle class imbalance in link prediction by defining a weighting factor $\zeta \in [0,1]$ to be used as a multiplier for the positive class in the cross-entropy loss formulation. For a single example $\mathbf{a}_i$ and its reconstructed output $\mathbf{\hat{a}}_i$, we compute the MBCE loss as follows:
\begin{align*}
\mathcal{L}_{\textsc{bce}} = -\mathbf{a}_i \log\left(\sigma\left(\mathbf{\hat{a}}_i\right)\right) \cdot \zeta - (1 - \mathbf{a}_i) &\log\left(1 - \sigma\left(\mathbf{\hat{a}}_i\right)\right), \\
\mathcal{L}_{\textsc{mbce}} = \frac{\sum_i \mathbf{m}_i \odot \mathcal{L}_{\textsc{bce}}} {\sum_i \mathbf{m}_i}.
\end{align*}
Here, $\mathcal{L}_{\textsc{bce}}$ is the balanced binary cross-entropy loss with weighting factor $\zeta = 1 - \frac{\text{\# positive links}}{\text{\# negative links}}$, $\sigma(\cdot)$ is the sigmoid function, $\odot$ is the Hadamard (element-wise) product,  and $\mathbf{m}_i$ is the Boolean mask: $\mathbf{m}_i = 1$  if $\mathbf{a}_i \neq \textsc{unk}$, else $\mathbf{m}_i = 0$.

\noindent \textbf{Semi-Supervised Node Classification} ~ The MTGAE model can also be used to perform efficient information propagation on graphs for the task of semi-supervised node classification. For a given augmented adjacency vector $\mathbf{\bar{a}}_i$, the model learns the corresponding node embedding $\mathbf{z}_i$ to obtain an optimal reconstruction. Intuitively, $\mathbf{z}_i$ encodes a vector of latent features derived from the concatenation of both graph and node features, and can be used to predict the label of the $i$th node. For multi-class classification, we decode $\mathbf{z}_i$ using the softmax activation function to produce a probability distribution over node labels. More precisely, we predict node labels via the following transformation: $\mathbf{\hat{y}}_i = \text{softmax}(\mathbf{\tilde{z}}_i) = \frac{1}{\mathcal{Z}} \exp(\mathbf{\tilde{z}}_i)$, where $\mathcal{Z} = \sum_i \exp(\mathbf{\tilde{z}}_i)$ and $\mathbf{\tilde{z}}_i = \mathbf{U} \boldsymbol{\cdot} \text{ReLU}\left(\mathbf{W}^\text{T} \mathbf{z}_i + \mathbf{b}^{(3)}\right) + \mathbf{b}^{(5)}$.

\noindent \textbf{Multi-Task Learning} ~ In many applications, such as knowledge base completion and network analysis, the input graph is partially observed with an incomplete set of edges and a small fraction of labeled nodes. Thus, it is desirable for a model to predict the labels of missing links and nodes \emph{simultaneously} in a multi-task learning setting. We achieve multi-task learning on graphs by training the MTGAE model using a joint loss function that combines the masked categorical cross-entropy loss for semi-supervised node classification with the MBCE loss for unsupervised link prediction:
\begin{equation*}
\mathcal{L}_{\textsc{multi-task}} = \overbrace{- \textsc{mask}_i \sum_{c \in C}\mathbf{y}_{ic} \log (\mathbf{\hat{y}}_{ic})}^\text{semi-supervised classification} + \mathcal{L}_{\textsc{mbce}},
\end{equation*}
where $C$ is the set of node labels, $\mathbf{y}_{ic}$ is the binary indicator if node $i$ belongs to class $c$, $\mathbf{\hat{y}}_{ic}$ is the softmax probability that node $i$ belongs to class $c$, $\mathcal{L}_{\textsc{mbce}}$ is the loss defined for unsupervised link prediction, and $\textsc{mask}_i$ is the Boolean variable: $\textsc{mask}_i = 1$ if node $i$ has a label, else $\textsc{mask}_i = 0$.

The training complexity of the MTGAE model is $\mathcal{O}((N+F)DI)$, where $N$ is the number of nodes, $F$ is the dimensionality of node features, $D$ is the size of the hidden layer, and $I$ is the number of iterations. In practice, $F$, $D \ll N$, and $I$ are independent of $N$. Thus, the overall complexity of MTGAE is $\mathcal{O}(N)$, linear in the number of nodes.

\section{Empirical Evaluation}

\begin{table}[ht]
\caption[Caption1]{Summary of datasets (\emph{left}) and baselines (\emph{right}) used in empirical evaluation. See \cite{Sen:2008,Wang:2016} for dataset details. The notation $\vert\mathit{O}^+\vert$:$\vert\mathit{O}^-\vert$ denotes the ratio of positive to negative edges and is a measure of class imbalance. Label rate is defined as the number of nodes labeled for training divided by the total number of nodes. Acronyms: AUC -- Area Under ROC Curve; AP -- Average Precision.}
\begin{minipage}[t]{0.5\textwidth}
\begin{center}
\begin{adjustbox}{width=0.95\textwidth}
	\begin{tabular} {l  r  r  r  r  r  r  r  r}
	\hline
	\multicolumn{1}{l}{\multirow{2}{*}{\textbf{Dataset}}} &
	\multicolumn{1}{c}{\multirow{2}{*}{\textbf{Nodes}}} &
    \multicolumn{1}{c}{\multirow{1}{*}{\textbf{Average}}} &
	\multicolumn{1}{c}{\multirow{1}{*}{\textbf{$\vert\mathit{O}^+\vert$:$\vert\mathit{O}^-\vert$}}} &
	\multicolumn{1}{c}{\multirow{1}{*}{\textbf{Node}}} &
	\multicolumn{1}{c}{\multirow{1}{*}{\textbf{Node}}} &
	\multicolumn{1}{c}{\multirow{1}{*}{\textbf{Label}}} \\
    & {}
    & \multicolumn{1}{c}{\multirow{1}{*}{\textbf{Degree}}}
    & \multicolumn{1}{c}{\multirow{1}{*}{\textbf{Ratio}}}
    & \multicolumn{1}{c}{\multirow{1}{*}{\textbf{Features}}}
    & \multicolumn{1}{c}{\multirow{1}{*}{\textbf{Classes}}}
    & \multicolumn{1}{c}{\multirow{1}{*}{\textbf{Rate}}} \\
    \hline \hline
    Pubmed
				& 19,717
				& 4.5
                & $1:4384$
				& 500
				& 3
				& 0.003 \\
    Citeseer
				& 3,327 
				& 2.8
                & $1:1198$
				& 3,703
				& 6
				& 0.036 \\
	Cora	
				& 2,708
				& 3.9
                & $1: \hspace{4.5pt} 694$
				& 1,433
				& 7
				& 0.052 \\
% 	Protein
% 					& 2,617
% 					& 9.1
%                     & $1: \hspace{4.5pt} 300$
% 					& 76
% 					& --
% 					& -- \\
% 	Metabolic
% 					& 668
% 					& 8.3
%                     & $1: \hspace{9pt} 80$
% 					& 325
% 					& --
% 					& -- \\
% 	Conflict
% 					& 130
% 					& 2.5
%                     & $1: \hspace{9pt} 52$
% 					& 3
% 					& --
% 					& -- \\
% 	PowerGrid
% 				 & 4,941
% 				 & 2.7
%                  & $1:1850$
% 				 & --
% 				 & --
% 				 & -- \\
    Arxiv-GRQC
    		    & 5,242
                & 5.5
                & $1: \hspace{4.5pt} 947$
                & --
                & --
                & -- \\
	BlogCatalog
    		    & 10,312
                & 64.8
                & $1: \hspace{4.5pt} 158$
                & --
                & --
                & -- \\
	\hline
	\end{tabular}
\end{adjustbox}
\end{center}
\end{minipage}
\begin{minipage}[t]{0.5\textwidth}
\begin{center}
\begin{adjustbox}{width=\textwidth}
	\begin{tabular} {l  l  l}
	\hline
	\multicolumn{1}{l}{\multirow{1}{*}{\textbf{Baseline}}} &
	\multicolumn{1}{l}{\multirow{1}{*}{\textbf{Evaluation Task}}} &
	\multicolumn{1}{l}{\multirow{1}{*}{\textbf{~~Metric}}} \\ \hline \hline
    SDNE \cite{Wang:2016} ~
							& Reconstruction
							& ~~Precision@$k$ \\
    % MF \cite{Menon:2011} ~
				% 	& Link Prediction
				% 	 & ~~AUC \\
    VGAE \cite{VGAE:2016} ~
				& Link Prediction
				 & ~~AUC, AP \\
    GCN \cite{Kipf:2016} ~
						& Node Classification
				 		& ~~Accuracy \\
	\hline
	\end{tabular}
\end{adjustbox}
\end{center}
\end{minipage}
\label{tab1}
\end{table}

\noindent \textbf{Implementation Details} ~ We closely follow the experimental protocols described in \cite{Kipf:2016,VGAE:2016} to train and evaluate our MTGAE model for LPNC. For link prediction, we form disjoint test, and validation, sets containing 10, and 5, percent of randomly sampled positive links and the same number of negative links, respectively, while utilizing all node features. For node classification, we split the data into disjoint test, and validation, sets of 1,000, and 500, examples, respectively and use only 20 examples per class for semi-supervised learning. In comparison to the baselines, we evaluate our MTGAE model on the same data splits over 10 runs with random weight initialization and report mean AUC/AP scores for link prediction and accuracy scores for node classification. We also compare the representation capacity of our MTGAE model against the related autoencoder-based SDNE model on the network reconstruction task. We use the ranking metric precision@$k$ to evaluate the model's ability to retrieve positive edges as part of the reconstruction.

Hyper-parameter tuning is performed on the validation set. Key hyper-parameters include mini-batch size, dimensionality of the hidden layers, and the percentage of dropout regularization \cite{Srivastava:2014}. In all experiments, the dimensionality of the hidden layers in the MTGAE architecture is fixed at $N$-256-128-256-$N$. We train for 100 epochs using Adam \cite{Kingma:2015} gradient descent with a fixed learning rate of $0.001$ on mini-batches of 64 examples.

We implement the MTGAE architecture using Keras \cite{Keras} on top of the GPU-enabled TensorFlow \cite{TF} backend. The diagonal elements of the adjacency matrix are set to $1$ with the interpretation that every node is connected to itself. We apply mean-variance normalization after each ReLU activation layer to help improve link prediction performance, where it compensates for noise between train and test instances by normalizing the activations to have zero mean and unit variance. During training, we implement several regularization techniques to mitigate overfitting, including dropout for highly sparse graphs and early stopping as a form of regularization in time when the model shows signs of overfitting on the validation set. We initialize weights according to the Glorot scheme described in \cite{Xavier:2010}. We do not apply weight decay regularization.

\begin{figure}[hb]
\centering
\includegraphics[width=0.88\textwidth]{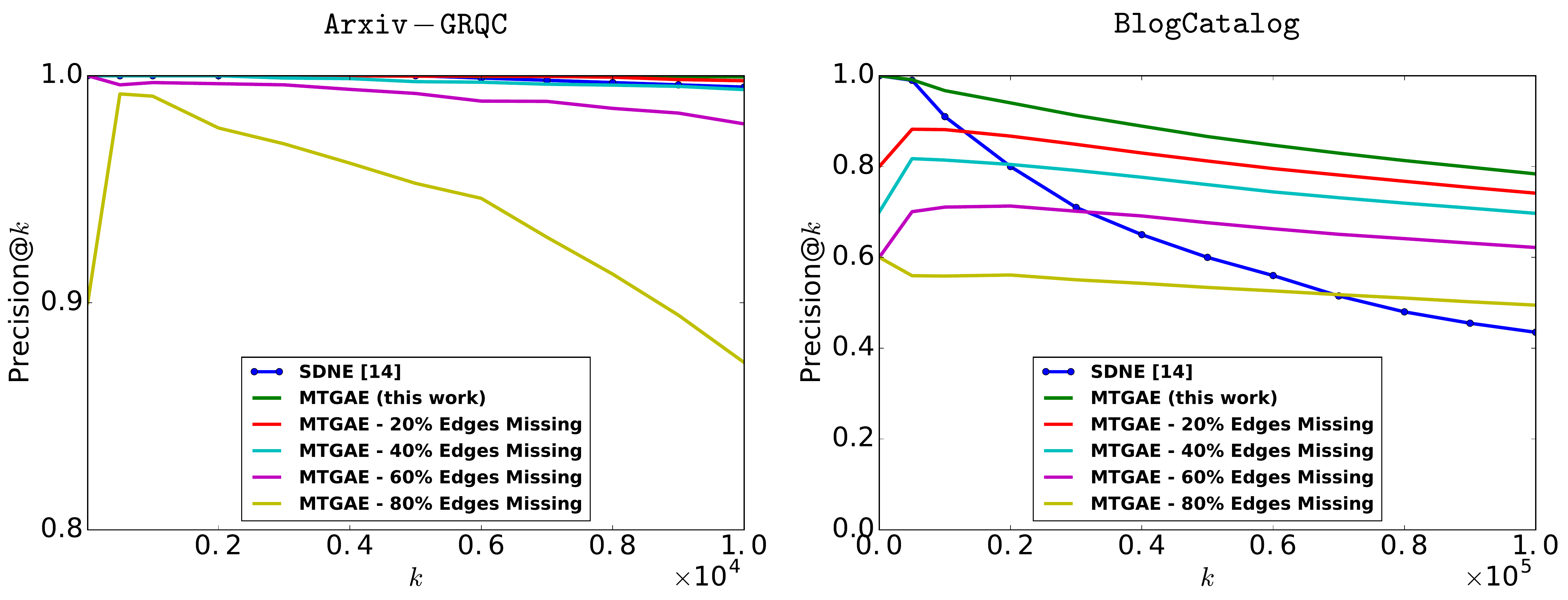}
\centering
\caption{Comparison of precision@$k$ performance between our MTGAE model and the related autoencoder-based SDNE model for the reconstruction task on the \texttt{Arxiv-GRQC} and \texttt{BlogCatalog} network datasets. The parameter $k$ indicates the total number of retrieved edges.}
\label{fig2}
\end{figure}

\noindent \textbf{Results and Analysis} ~ Results of the reconstruction task for the \texttt{Arxiv-GRQC} and \texttt{BlogCatalog} network datasets are illustrated in Figure~\ref{fig2}. In comparison to SDNE, we show that our MTGAE model achieves better precision@$k$ performance for all $k$ values, up to $k=10,000$ for \texttt{Arxiv-GRQC} and $k=100,000$ for \texttt{BlogCatalog}, when trained on the complete datasets. We also systematically test the capacity of the MTGAE model to reconstruct the original networks when up to 80 percent of the edges are randomly removed, akin to the link prediction task. We show that the MTGAE model only gets worse precision@$k$ performance than SDNE on the \texttt{Arxiv-GRQC} dataset when more than 40 percent of the edges are missing. On the \texttt{BlogCatalog} dataset, the MTGAE model achieves better precision@$k$ performance than SDNE for large $k$ values even when 80 percent of the edges are missing at random. This experiment demonstrates the superior representation capacity of our MTGAE model when compared to SDNE, which is attributed to parameter sharing in the architecture.

Lastly, we report LPNC results obtained by our MTGAE model in the MTL scenario. The model takes as input an incomplete graph with 10 percent of the positive edges, and the same number of negative edges, missing at random and all available node features to simultaneously predict labels for the nodes and missing edges. Table~\ref{tab2} shows the efficacy of the MTGAE model for MTL when compared against recent state-of-the-art \emph{task-specific} link prediction and node classification models, which require the complete adjacency matrix as input. For link prediction, MTGAE significantly outperforms the best VGAE model on \texttt{Cora} and \texttt{Citeseer}. For node classification, MTGAE is the best performing model on the \texttt{Citeseer} and \texttt{Pubmed} datasets, which have very low node label rates.

\begin{table}[ht]
\begin{center}
\caption[Caption2]{Comparison of LPNC performances between our MTGAE model and recent state-of-the-art graph embedding methods. Link prediction performance is reported as the combined average of AUC and AP scores. Accuracy is used for node classification performance.}
\begin{adjustbox}{width=0.5\textwidth}
	\begin{tabular} {l  r  r  r}
	\hline
	\multicolumn{1}{l}{\multirow{1}{*}{\textbf{Method}} } &
	\multicolumn{1}{r}{\multirow{1}{*}{\textbf{Cora}}} &
	\multicolumn{1}{r}{\multirow{1}{*}{\textbf{~Citeseer}}} &
	\multicolumn{1}{r}{\multirow{1}{*}{\textbf{~Pubmed}}} \\ \hline \hline
    \multicolumn{4}{c}{\multirow{1}{*}{\textbf{Link Prediction}}} \\
	MTGAE ~
						& \textbf{0.946}
						& \textbf{0.949}
						& 0.944 \\
	VGAE \cite{VGAE:2016} ~
				 & 0.920
				 & 0.914
				 & \textbf{0.965} \\
	\hline \hline
    \multicolumn{4}{c}{\multirow{1}{*}{\textbf{Node Classification}}} \\
	MTGAE
							& 0.790
							& \textbf{0.718}
							&  \textbf{0.804} \\
	GCN \cite{Kipf:2016} 
					 & \textbf{0.815}
					 & 0.703
					 & 0.790 \\
	Planetoid \cite{Yang:2016}
	                & 0.757
	                & 0.647
	                & 0.772 \\
	\hline
	\end{tabular}
	\label{tab2}
\end{adjustbox}
\end{center}
\end{table}

\noindent \textbf{Future Work} ~ Further research will explore inductive reasoning on out-of-network nodes and mitigate $\mathcal{O}(N)$ complexity for improved scalability on large, dynamic graphs.

\newpage


\end{document}